\documentclass{article}

\usepackage[preprint]{neurips_2019}

\usepackage[utf8]{inputenc} 
\usepackage[T1]{fontenc}    
\usepackage{hyperref}       
\usepackage{url}            
\usepackage{booktabs}       
\usepackage{amsfonts}       
\usepackage{nicefrac}       
\usepackage{microtype}      

\title{\emph{TorchGAN}: A Flexible Framework for GAN Training and Evaluation}

\usepackage{makecell}
\usepackage{mydefs}
\usepackage{notes}
\usepackage{url}
\usepackage{tikz}
\usepackage{lipsum}
\usepackage[labelfont=bf]{caption}
\usepackage{subcaption}
\usepackage{amsmath}
\usepackage[linesnumbered, vlined, ruled]{algorithm2e}
\usepackage{xpatch}
\usepackage{todonotes}
\usepackage{booktabs}
\usepackage[draft]{minted}
\usepackage{xcolor}
\usepackage{hyperref}
\usepackage{color}
\hypersetup{colorlinks,linkcolor={blue},citecolor={green},urlcolor={blue}}

\makeatletter
\xpatchcmd{\@thm}{\thm@headpunct{.}}{\thm@headpunct{}}{}{}
\makeatother

\theoremstyle{definition}

\makeatletter
\def\PYG@reset{\let\PYG@it=\relax \let\PYG@bf=\relax%
    \let\PYG@ul=\relax \let\PYG@tc=\relax%
    \let\PYG@bc=\relax \let\PYG@ff=\relax}
\def\PYG@tok#1{\csname PYG@tok@#1\endcsname}
\def\PYG@toks#1+{\ifx\relax#1\empty\else%
    \PYG@tok{#1}\expandafter\PYG@toks\fi}
\def\PYG@do#1{\PYG@bc{\PYG@tc{\PYG@ul{%
    \PYG@it{\PYG@bf{\PYG@ff{#1}}}}}}}
\def\PYG#1#2{\PYG@reset\PYG@toks#1+\relax+\PYG@do{#2}}

\expandafter\def\csname PYG@tok@w\endcsname{\def\PYG@tc##1{\textcolor[rgb]{0.73,0.73,0.73}{##1}}}
\expandafter\def\csname PYG@tok@c\endcsname{\let\PYG@it=\textit\def\PYG@tc##1{\textcolor[rgb]{0.25,0.50,0.50}{##1}}}
\expandafter\def\csname PYG@tok@cp\endcsname{\def\PYG@tc##1{\textcolor[rgb]{0.74,0.48,0.00}{##1}}}
\expandafter\def\csname PYG@tok@k\endcsname{\let\PYG@bf=\textbf\def\PYG@tc##1{\textcolor[rgb]{0.00,0.50,0.00}{##1}}}
\expandafter\def\csname PYG@tok@kp\endcsname{\def\PYG@tc##1{\textcolor[rgb]{0.00,0.50,0.00}{##1}}}
\expandafter\def\csname PYG@tok@kt\endcsname{\def\PYG@tc##1{\textcolor[rgb]{0.69,0.00,0.25}{##1}}}
\expandafter\def\csname PYG@tok@o\endcsname{\def\PYG@tc##1{\textcolor[rgb]{0.40,0.40,0.40}{##1}}}
\expandafter\def\csname PYG@tok@ow\endcsname{\let\PYG@bf=\textbf\def\PYG@tc##1{\textcolor[rgb]{0.67,0.13,1.00}{##1}}}
\expandafter\def\csname PYG@tok@nb\endcsname{\def\PYG@tc##1{\textcolor[rgb]{0.00,0.50,0.00}{##1}}}
\expandafter\def\csname PYG@tok@nf\endcsname{\def\PYG@tc##1{\textcolor[rgb]{0.00,0.00,1.00}{##1}}}
\expandafter\def\csname PYG@tok@nc\endcsname{\let\PYG@bf=\textbf\def\PYG@tc##1{\textcolor[rgb]{0.00,0.00,1.00}{##1}}}
\expandafter\def\csname PYG@tok@nn\endcsname{\let\PYG@bf=\textbf\def\PYG@tc##1{\textcolor[rgb]{0.00,0.00,1.00}{##1}}}
\expandafter\def\csname PYG@tok@ne\endcsname{\let\PYG@bf=\textbf\def\PYG@tc##1{\textcolor[rgb]{0.82,0.25,0.23}{##1}}}
\expandafter\def\csname PYG@tok@nv\endcsname{\def\PYG@tc##1{\textcolor[rgb]{0.10,0.09,0.49}{##1}}}
\expandafter\def\csname PYG@tok@no\endcsname{\def\PYG@tc##1{\textcolor[rgb]{0.53,0.00,0.00}{##1}}}
\expandafter\def\csname PYG@tok@nl\endcsname{\def\PYG@tc##1{\textcolor[rgb]{0.63,0.63,0.00}{##1}}}
\expandafter\def\csname PYG@tok@ni\endcsname{\let\PYG@bf=\textbf\def\PYG@tc##1{\textcolor[rgb]{0.60,0.60,0.60}{##1}}}
\expandafter\def\csname PYG@tok@na\endcsname{\def\PYG@tc##1{\textcolor[rgb]{0.49,0.56,0.16}{##1}}}
\expandafter\def\csname PYG@tok@nt\endcsname{\let\PYG@bf=\textbf\def\PYG@tc##1{\textcolor[rgb]{0.00,0.50,0.00}{##1}}}
\expandafter\def\csname PYG@tok@nd\endcsname{\def\PYG@tc##1{\textcolor[rgb]{0.67,0.13,1.00}{##1}}}
\expandafter\def\csname PYG@tok@s\endcsname{\def\PYG@tc##1{\textcolor[rgb]{0.73,0.13,0.13}{##1}}}
\expandafter\def\csname PYG@tok@sd\endcsname{\let\PYG@it=\textit\def\PYG@tc##1{\textcolor[rgb]{0.73,0.13,0.13}{##1}}}
\expandafter\def\csname PYG@tok@si\endcsname{\let\PYG@bf=\textbf\def\PYG@tc##1{\textcolor[rgb]{0.73,0.40,0.53}{##1}}}
\expandafter\def\csname PYG@tok@se\endcsname{\let\PYG@bf=\textbf\def\PYG@tc##1{\textcolor[rgb]{0.73,0.40,0.13}{##1}}}
\expandafter\def\csname PYG@tok@sr\endcsname{\def\PYG@tc##1{\textcolor[rgb]{0.73,0.40,0.53}{##1}}}
\expandafter\def\csname PYG@tok@ss\endcsname{\def\PYG@tc##1{\textcolor[rgb]{0.10,0.09,0.49}{##1}}}
\expandafter\def\csname PYG@tok@sx\endcsname{\def\PYG@tc##1{\textcolor[rgb]{0.00,0.50,0.00}{##1}}}
\expandafter\def\csname PYG@tok@m\endcsname{\def\PYG@tc##1{\textcolor[rgb]{0.40,0.40,0.40}{##1}}}
\expandafter\def\csname PYG@tok@gh\endcsname{\let\PYG@bf=\textbf\def\PYG@tc##1{\textcolor[rgb]{0.00,0.00,0.50}{##1}}}
\expandafter\def\csname PYG@tok@gu\endcsname{\let\PYG@bf=\textbf\def\PYG@tc##1{\textcolor[rgb]{0.50,0.00,0.50}{##1}}}
\expandafter\def\csname PYG@tok@gd\endcsname{\def\PYG@tc##1{\textcolor[rgb]{0.63,0.00,0.00}{##1}}}
\expandafter\def\csname PYG@tok@gi\endcsname{\def\PYG@tc##1{\textcolor[rgb]{0.00,0.63,0.00}{##1}}}
\expandafter\def\csname PYG@tok@gr\endcsname{\def\PYG@tc##1{\textcolor[rgb]{1.00,0.00,0.00}{##1}}}
\expandafter\def\csname PYG@tok@ge\endcsname{\let\PYG@it=\textit}
\expandafter\def\csname PYG@tok@gs\endcsname{\let\PYG@bf=\textbf}
\expandafter\def\csname PYG@tok@gp\endcsname{\let\PYG@bf=\textbf\def\PYG@tc##1{\textcolor[rgb]{0.00,0.00,0.50}{##1}}}
\expandafter\def\csname PYG@tok@go\endcsname{\def\PYG@tc##1{\textcolor[rgb]{0.53,0.53,0.53}{##1}}}
\expandafter\def\csname PYG@tok@gt\endcsname{\def\PYG@tc##1{\textcolor[rgb]{0.00,0.27,0.87}{##1}}}
\expandafter\def\csname PYG@tok@err\endcsname{\def\PYG@bc##1{\setlength{\fboxsep}{0pt}\fcolorbox[rgb]{1.00,0.00,0.00}{1,1,1}{\strut ##1}}}
\expandafter\def\csname PYG@tok@kc\endcsname{\let\PYG@bf=\textbf\def\PYG@tc##1{\textcolor[rgb]{0.00,0.50,0.00}{##1}}}
\expandafter\def\csname PYG@tok@kd\endcsname{\let\PYG@bf=\textbf\def\PYG@tc##1{\textcolor[rgb]{0.00,0.50,0.00}{##1}}}
\expandafter\def\csname PYG@tok@kn\endcsname{\let\PYG@bf=\textbf\def\PYG@tc##1{\textcolor[rgb]{0.00,0.50,0.00}{##1}}}
\expandafter\def\csname PYG@tok@kr\endcsname{\let\PYG@bf=\textbf\def\PYG@tc##1{\textcolor[rgb]{0.00,0.50,0.00}{##1}}}
\expandafter\def\csname PYG@tok@bp\endcsname{\def\PYG@tc##1{\textcolor[rgb]{0.00,0.50,0.00}{##1}}}
\expandafter\def\csname PYG@tok@fm\endcsname{\def\PYG@tc##1{\textcolor[rgb]{0.00,0.00,1.00}{##1}}}
\expandafter\def\csname PYG@tok@vc\endcsname{\def\PYG@tc##1{\textcolor[rgb]{0.10,0.09,0.49}{##1}}}
\expandafter\def\csname PYG@tok@vg\endcsname{\def\PYG@tc##1{\textcolor[rgb]{0.10,0.09,0.49}{##1}}}
\expandafter\def\csname PYG@tok@vi\endcsname{\def\PYG@tc##1{\textcolor[rgb]{0.10,0.09,0.49}{##1}}}
\expandafter\def\csname PYG@tok@vm\endcsname{\def\PYG@tc##1{\textcolor[rgb]{0.10,0.09,0.49}{##1}}}
\expandafter\def\csname PYG@tok@sa\endcsname{\def\PYG@tc##1{\textcolor[rgb]{0.73,0.13,0.13}{##1}}}
\expandafter\def\csname PYG@tok@sb\endcsname{\def\PYG@tc##1{\textcolor[rgb]{0.73,0.13,0.13}{##1}}}
\expandafter\def\csname PYG@tok@sc\endcsname{\def\PYG@tc##1{\textcolor[rgb]{0.73,0.13,0.13}{##1}}}
\expandafter\def\csname PYG@tok@dl\endcsname{\def\PYG@tc##1{\textcolor[rgb]{0.73,0.13,0.13}{##1}}}
\expandafter\def\csname PYG@tok@s2\endcsname{\def\PYG@tc##1{\textcolor[rgb]{0.73,0.13,0.13}{##1}}}
\expandafter\def\csname PYG@tok@sh\endcsname{\def\PYG@tc##1{\textcolor[rgb]{0.73,0.13,0.13}{##1}}}
\expandafter\def\csname PYG@tok@s1\endcsname{\def\PYG@tc##1{\textcolor[rgb]{0.73,0.13,0.13}{##1}}}
\expandafter\def\csname PYG@tok@mb\endcsname{\def\PYG@tc##1{\textcolor[rgb]{0.40,0.40,0.40}{##1}}}
\expandafter\def\csname PYG@tok@mf\endcsname{\def\PYG@tc##1{\textcolor[rgb]{0.40,0.40,0.40}{##1}}}
\expandafter\def\csname PYG@tok@mh\endcsname{\def\PYG@tc##1{\textcolor[rgb]{0.40,0.40,0.40}{##1}}}
\expandafter\def\csname PYG@tok@mi\endcsname{\def\PYG@tc##1{\textcolor[rgb]{0.40,0.40,0.40}{##1}}}
\expandafter\def\csname PYG@tok@il\endcsname{\def\PYG@tc##1{\textcolor[rgb]{0.40,0.40,0.40}{##1}}}
\expandafter\def\csname PYG@tok@mo\endcsname{\def\PYG@tc##1{\textcolor[rgb]{0.40,0.40,0.40}{##1}}}
\expandafter\def\csname PYG@tok@ch\endcsname{\let\PYG@it=\textit\def\PYG@tc##1{\textcolor[rgb]{0.25,0.50,0.50}{##1}}}
\expandafter\def\csname PYG@tok@cm\endcsname{\let\PYG@it=\textit\def\PYG@tc##1{\textcolor[rgb]{0.25,0.50,0.50}{##1}}}
\expandafter\def\csname PYG@tok@cpf\endcsname{\let\PYG@it=\textit\def\PYG@tc##1{\textcolor[rgb]{0.25,0.50,0.50}{##1}}}
\expandafter\def\csname PYG@tok@c1\endcsname{\let\PYG@it=\textit\def\PYG@tc##1{\textcolor[rgb]{0.25,0.50,0.50}{##1}}}
\expandafter\def\csname PYG@tok@cs\endcsname{\let\PYG@it=\textit\def\PYG@tc##1{\textcolor[rgb]{0.25,0.50,0.50}{##1}}}

\makeatother

\makeatletter
\def\PYGdefault@reset{\let\PYGdefault@it=\relax \let\PYGdefault@bf=\relax%
    \let\PYGdefault@ul=\relax \let\PYGdefault@tc=\relax%
    \let\PYGdefault@bc=\relax \let\PYGdefault@ff=\relax}
\def\PYGdefault@tok#1{\csname PYGdefault@tok@#1\endcsname}
\def\PYGdefault@toks#1+{\ifx\relax#1\empty\else%
    \PYGdefault@tok{#1}\expandafter\PYGdefault@toks\fi}
\def\PYGdefault@do#1{\PYGdefault@bc{\PYGdefault@tc{\PYGdefault@ul{%
    \PYGdefault@it{\PYGdefault@bf{\PYGdefault@ff{#1}}}}}}}
\def\PYGdefault#1#2{\PYGdefault@reset\PYGdefault@toks#1+\relax+\PYGdefault@do{#2}}

\expandafter\def\csname PYGdefault@tok@w\endcsname{\def\PYGdefault@tc##1{\textcolor[rgb]{0.73,0.73,0.73}{##1}}}
\expandafter\def\csname PYGdefault@tok@c\endcsname{\let\PYGdefault@it=\textit\def\PYGdefault@tc##1{\textcolor[rgb]{0.25,0.50,0.50}{##1}}}
\expandafter\def\csname PYGdefault@tok@cp\endcsname{\def\PYGdefault@tc##1{\textcolor[rgb]{0.74,0.48,0.00}{##1}}}
\expandafter\def\csname PYGdefault@tok@k\endcsname{\let\PYGdefault@bf=\textbf\def\PYGdefault@tc##1{\textcolor[rgb]{0.00,0.50,0.00}{##1}}}
\expandafter\def\csname PYGdefault@tok@kp\endcsname{\def\PYGdefault@tc##1{\textcolor[rgb]{0.00,0.50,0.00}{##1}}}
\expandafter\def\csname PYGdefault@tok@kt\endcsname{\def\PYGdefault@tc##1{\textcolor[rgb]{0.69,0.00,0.25}{##1}}}
\expandafter\def\csname PYGdefault@tok@o\endcsname{\def\PYGdefault@tc##1{\textcolor[rgb]{0.40,0.40,0.40}{##1}}}
\expandafter\def\csname PYGdefault@tok@ow\endcsname{\let\PYGdefault@bf=\textbf\def\PYGdefault@tc##1{\textcolor[rgb]{0.67,0.13,1.00}{##1}}}
\expandafter\def\csname PYGdefault@tok@nb\endcsname{\def\PYGdefault@tc##1{\textcolor[rgb]{0.00,0.50,0.00}{##1}}}
\expandafter\def\csname PYGdefault@tok@nf\endcsname{\def\PYGdefault@tc##1{\textcolor[rgb]{0.00,0.00,1.00}{##1}}}
\expandafter\def\csname PYGdefault@tok@nc\endcsname{\let\PYGdefault@bf=\textbf\def\PYGdefault@tc##1{\textcolor[rgb]{0.00,0.00,1.00}{##1}}}
\expandafter\def\csname PYGdefault@tok@nn\endcsname{\let\PYGdefault@bf=\textbf\def\PYGdefault@tc##1{\textcolor[rgb]{0.00,0.00,1.00}{##1}}}
\expandafter\def\csname PYGdefault@tok@ne\endcsname{\let\PYGdefault@bf=\textbf\def\PYGdefault@tc##1{\textcolor[rgb]{0.82,0.25,0.23}{##1}}}
\expandafter\def\csname PYGdefault@tok@nv\endcsname{\def\PYGdefault@tc##1{\textcolor[rgb]{0.10,0.09,0.49}{##1}}}
\expandafter\def\csname PYGdefault@tok@no\endcsname{\def\PYGdefault@tc##1{\textcolor[rgb]{0.53,0.00,0.00}{##1}}}
\expandafter\def\csname PYGdefault@tok@nl\endcsname{\def\PYGdefault@tc##1{\textcolor[rgb]{0.63,0.63,0.00}{##1}}}
\expandafter\def\csname PYGdefault@tok@ni\endcsname{\let\PYGdefault@bf=\textbf\def\PYGdefault@tc##1{\textcolor[rgb]{0.60,0.60,0.60}{##1}}}
\expandafter\def\csname PYGdefault@tok@na\endcsname{\def\PYGdefault@tc##1{\textcolor[rgb]{0.49,0.56,0.16}{##1}}}
\expandafter\def\csname PYGdefault@tok@nt\endcsname{\let\PYGdefault@bf=\textbf\def\PYGdefault@tc##1{\textcolor[rgb]{0.00,0.50,0.00}{##1}}}
\expandafter\def\csname PYGdefault@tok@nd\endcsname{\def\PYGdefault@tc##1{\textcolor[rgb]{0.67,0.13,1.00}{##1}}}
\expandafter\def\csname PYGdefault@tok@s\endcsname{\def\PYGdefault@tc##1{\textcolor[rgb]{0.73,0.13,0.13}{##1}}}
\expandafter\def\csname PYGdefault@tok@sd\endcsname{\let\PYGdefault@it=\textit\def\PYGdefault@tc##1{\textcolor[rgb]{0.73,0.13,0.13}{##1}}}
\expandafter\def\csname PYGdefault@tok@si\endcsname{\let\PYGdefault@bf=\textbf\def\PYGdefault@tc##1{\textcolor[rgb]{0.73,0.40,0.53}{##1}}}
\expandafter\def\csname PYGdefault@tok@se\endcsname{\let\PYGdefault@bf=\textbf\def\PYGdefault@tc##1{\textcolor[rgb]{0.73,0.40,0.13}{##1}}}
\expandafter\def\csname PYGdefault@tok@sr\endcsname{\def\PYGdefault@tc##1{\textcolor[rgb]{0.73,0.40,0.53}{##1}}}
\expandafter\def\csname PYGdefault@tok@ss\endcsname{\def\PYGdefault@tc##1{\textcolor[rgb]{0.10,0.09,0.49}{##1}}}
\expandafter\def\csname PYGdefault@tok@sx\endcsname{\def\PYGdefault@tc##1{\textcolor[rgb]{0.00,0.50,0.00}{##1}}}
\expandafter\def\csname PYGdefault@tok@m\endcsname{\def\PYGdefault@tc##1{\textcolor[rgb]{0.40,0.40,0.40}{##1}}}
\expandafter\def\csname PYGdefault@tok@gh\endcsname{\let\PYGdefault@bf=\textbf\def\PYGdefault@tc##1{\textcolor[rgb]{0.00,0.00,0.50}{##1}}}
\expandafter\def\csname PYGdefault@tok@gu\endcsname{\let\PYGdefault@bf=\textbf\def\PYGdefault@tc##1{\textcolor[rgb]{0.50,0.00,0.50}{##1}}}
\expandafter\def\csname PYGdefault@tok@gd\endcsname{\def\PYGdefault@tc##1{\textcolor[rgb]{0.63,0.00,0.00}{##1}}}
\expandafter\def\csname PYGdefault@tok@gi\endcsname{\def\PYGdefault@tc##1{\textcolor[rgb]{0.00,0.63,0.00}{##1}}}
\expandafter\def\csname PYGdefault@tok@gr\endcsname{\def\PYGdefault@tc##1{\textcolor[rgb]{1.00,0.00,0.00}{##1}}}
\expandafter\def\csname PYGdefault@tok@ge\endcsname{\let\PYGdefault@it=\textit}
\expandafter\def\csname PYGdefault@tok@gs\endcsname{\let\PYGdefault@bf=\textbf}
\expandafter\def\csname PYGdefault@tok@gp\endcsname{\let\PYGdefault@bf=\textbf\def\PYGdefault@tc##1{\textcolor[rgb]{0.00,0.00,0.50}{##1}}}
\expandafter\def\csname PYGdefault@tok@go\endcsname{\def\PYGdefault@tc##1{\textcolor[rgb]{0.53,0.53,0.53}{##1}}}
\expandafter\def\csname PYGdefault@tok@gt\endcsname{\def\PYGdefault@tc##1{\textcolor[rgb]{0.00,0.27,0.87}{##1}}}
\expandafter\def\csname PYGdefault@tok@err\endcsname{\def\PYGdefault@bc##1{\setlength{\fboxsep}{0pt}\fcolorbox[rgb]{1.00,0.00,0.00}{1,1,1}{\strut ##1}}}
\expandafter\def\csname PYGdefault@tok@kc\endcsname{\let\PYGdefault@bf=\textbf\def\PYGdefault@tc##1{\textcolor[rgb]{0.00,0.50,0.00}{##1}}}
\expandafter\def\csname PYGdefault@tok@kd\endcsname{\let\PYGdefault@bf=\textbf\def\PYGdefault@tc##1{\textcolor[rgb]{0.00,0.50,0.00}{##1}}}
\expandafter\def\csname PYGdefault@tok@kn\endcsname{\let\PYGdefault@bf=\textbf\def\PYGdefault@tc##1{\textcolor[rgb]{0.00,0.50,0.00}{##1}}}
\expandafter\def\csname PYGdefault@tok@kr\endcsname{\let\PYGdefault@bf=\textbf\def\PYGdefault@tc##1{\textcolor[rgb]{0.00,0.50,0.00}{##1}}}
\expandafter\def\csname PYGdefault@tok@bp\endcsname{\def\PYGdefault@tc##1{\textcolor[rgb]{0.00,0.50,0.00}{##1}}}
\expandafter\def\csname PYGdefault@tok@fm\endcsname{\def\PYGdefault@tc##1{\textcolor[rgb]{0.00,0.00,1.00}{##1}}}
\expandafter\def\csname PYGdefault@tok@vc\endcsname{\def\PYGdefault@tc##1{\textcolor[rgb]{0.10,0.09,0.49}{##1}}}
\expandafter\def\csname PYGdefault@tok@vg\endcsname{\def\PYGdefault@tc##1{\textcolor[rgb]{0.10,0.09,0.49}{##1}}}
\expandafter\def\csname PYGdefault@tok@vi\endcsname{\def\PYGdefault@tc##1{\textcolor[rgb]{0.10,0.09,0.49}{##1}}}
\expandafter\def\csname PYGdefault@tok@vm\endcsname{\def\PYGdefault@tc##1{\textcolor[rgb]{0.10,0.09,0.49}{##1}}}
\expandafter\def\csname PYGdefault@tok@sa\endcsname{\def\PYGdefault@tc##1{\textcolor[rgb]{0.73,0.13,0.13}{##1}}}
\expandafter\def\csname PYGdefault@tok@sb\endcsname{\def\PYGdefault@tc##1{\textcolor[rgb]{0.73,0.13,0.13}{##1}}}
\expandafter\def\csname PYGdefault@tok@sc\endcsname{\def\PYGdefault@tc##1{\textcolor[rgb]{0.73,0.13,0.13}{##1}}}
\expandafter\def\csname PYGdefault@tok@dl\endcsname{\def\PYGdefault@tc##1{\textcolor[rgb]{0.73,0.13,0.13}{##1}}}
\expandafter\def\csname PYGdefault@tok@s2\endcsname{\def\PYGdefault@tc##1{\textcolor[rgb]{0.73,0.13,0.13}{##1}}}
\expandafter\def\csname PYGdefault@tok@sh\endcsname{\def\PYGdefault@tc##1{\textcolor[rgb]{0.73,0.13,0.13}{##1}}}
\expandafter\def\csname PYGdefault@tok@s1\endcsname{\def\PYGdefault@tc##1{\textcolor[rgb]{0.73,0.13,0.13}{##1}}}
\expandafter\def\csname PYGdefault@tok@mb\endcsname{\def\PYGdefault@tc##1{\textcolor[rgb]{0.40,0.40,0.40}{##1}}}
\expandafter\def\csname PYGdefault@tok@mf\endcsname{\def\PYGdefault@tc##1{\textcolor[rgb]{0.40,0.40,0.40}{##1}}}
\expandafter\def\csname PYGdefault@tok@mh\endcsname{\def\PYGdefault@tc##1{\textcolor[rgb]{0.40,0.40,0.40}{##1}}}
\expandafter\def\csname PYGdefault@tok@mi\endcsname{\def\PYGdefault@tc##1{\textcolor[rgb]{0.40,0.40,0.40}{##1}}}
\expandafter\def\csname PYGdefault@tok@il\endcsname{\def\PYGdefault@tc##1{\textcolor[rgb]{0.40,0.40,0.40}{##1}}}
\expandafter\def\csname PYGdefault@tok@mo\endcsname{\def\PYGdefault@tc##1{\textcolor[rgb]{0.40,0.40,0.40}{##1}}}
\expandafter\def\csname PYGdefault@tok@ch\endcsname{\let\PYGdefault@it=\textit\def\PYGdefault@tc##1{\textcolor[rgb]{0.25,0.50,0.50}{##1}}}
\expandafter\def\csname PYGdefault@tok@cm\endcsname{\let\PYGdefault@it=\textit\def\PYGdefault@tc##1{\textcolor[rgb]{0.25,0.50,0.50}{##1}}}
\expandafter\def\csname PYGdefault@tok@cpf\endcsname{\let\PYGdefault@it=\textit\def\PYGdefault@tc##1{\textcolor[rgb]{0.25,0.50,0.50}{##1}}}
\expandafter\def\csname PYGdefault@tok@c1\endcsname{\let\PYGdefault@it=\textit\def\PYGdefault@tc##1{\textcolor[rgb]{0.25,0.50,0.50}{##1}}}
\expandafter\def\csname PYGdefault@tok@cs\endcsname{\let\PYGdefault@it=\textit\def\PYGdefault@tc##1{\textcolor[rgb]{0.25,0.50,0.50}{##1}}}

\makeatother

\author{%
    Avik Pal*\\
    avik.pal.2017@gmail.com\\
    Department of Computer Science and Engineering\\
    Indian Institute of Technology Kanpur\\
    \And
    Aniket Das*\\
    aniketd@iitk.ac.in\\
    Department of Electrical Engineering\\
    Indian Institute of Technology Kanpur\\
}

\newcommand{\describeContent}[1]{%
    \begingroup%
    \let\thefootnote\relax%
    \footnotetext{#1}%
    \endgroup%
}

\begin{document}

\maketitle

\begin{abstract}
TorchGAN is a PyTorch based framework for writing succinct and comprehensible code for training and evaluation of Generative Adversarial Networks. The framework's modular design allows effortless customization of the model architecture, loss functions, training paradigms, and evaluation metrics. The key features of TorchGAN are its extensibility, built-in support for a large number of popular models, losses and evaluation metrics, and zero overhead compared to vanilla PyTorch. By using the framework to implement several popular GAN models, we demonstrate its extensibility and ease of use. We also benchmark the training time of our framework for said models against the corresponding baseline PyTorch implementations and observe that TorchGAN's features bear almost zero overhead.
\end{abstract}

\describeContent{* Equal Contribution}

\section{Introduction}

Generative Adversarial Networks (GANs) (\cite{gan2014}) are a class of deep generative models that formulate the model estimation problem as an adversarial game between two neural networks, a Generator representing an implicit generative distribution, and a Discriminator that differentiates between samples from said implicit distribution and the true data distribution. The implicit distribution recovers the data distribution when the game reaches equilibrium. Apart from being one of the most popular approaches for generative modeling and unsupervised learning tasks in Computer Vision, with diverse applications such as photo-realistic image generation (\cite{biggan}, \cite{karras2017progressive}), image super-resolution (\cite{srgan}), image-to-image translation (\cite{cyclegan}) and video generation (\cite{dvdgan}, \cite{mocogan}), it has also found applicability in domains such as Natural Language Processing (\cite{zhang2017adversarial}) and Time Series Analysis (\cite{esteban2017real}).

GANs generally share a standard design paradigm, with the building blocks comprising one or more generator and discriminator models, and the associated loss functions for training them. TorchGAN makes use of this design similarity by exposing a simple API for customizing these blocks. The interaction between these components at training time is facilitated by a highly robust trainer which automatically adapts to user-defined GAN models and losses. TorchGAN provides an extensive and continually expanding collection of popular GAN models, losses, evaluation metrics, and stability-enhancing features, which can either be used off the shelf or easily extended or combined to design more sophisticated models effortlessly. With the above design principles in mind, we aim to improve upon existing GAN training frameworks such as TFGAN~\cite{tfgan}, HyperGAN~\cite{hypergan}, and IBM GAN-Toolkit~\cite{gantoolkit} on the aspects of extensibility, the richness of the feature set and documentation.

\begin{figure}[!htb]
    \centering
\begin{minted}{python}
train_dataset = dsets.CIFAR10(root='./cifar10', train=True, transform=transforms.Compose([transforms.ToTensor(), transforms.Normalize(mean=(0.5, 0.5, 0.5), std=(0.5, 0.5, 0.5))]), download=True)
train_loader = data.DataLoader(train_dataset, batch_size=128, shuffle=True)
trainer = Trainer(
    {"generator": {"name": DCGANGenerator, "args": {"out_channels": 3, "step_channels": 16}, "optimizer": {"name": Adam, "args": {"lr": 0.0002, "betas": (0.5, 0.999)}}}, 
    "discriminator": {"name": DCGANDiscriminator, "args": {"in_channels": 3, "step_channels": 16}, "optimizer": {"name": Adam, "args": {"lr": 0.0002, "betas": (0.5, 0.999)}}}}, [MinimaxGeneratorLoss(), MinimaxDiscriminatorLoss()], sample_size=64, epochs=20)
trainer(train_loader)
\end{minted}
    \caption{DCGAN (\cite{dcgan2015}) in under 10 lines of code with TorchGAN}
    \label{fig:get_started}
\end{figure}

\section{Implementing Models in TorchGAN}

The core of the TorchGAN framework is a highly versatile trainer module, responsible for its flexibility and ease of use. The trainer requires specification of the generator and the discriminator architecture along with the optimizers associated with each of them, represented as a dictionary, as well as the list of associated loss functions, and optionally, evaluation metrics.  We provide an illustrative example for training DCGAN on CIFAR10 in Figure~\ref{fig:get_started}. One can either choose from the in-built implementations of popular GAN models, losses and metrics or define custom variants of their own with minimal effort by extending the appropriate base classes. This extensibility is widely useful in research applications where the user only needs to write code for the model architecture and/or the loss function. The trainer automatically handles the intricacies of training with custom models/losses. The trainer also supports the usage of multiple generators and discriminators, allowing training of more sophisticated models such as Generative Multi Adversarial Networks (\cite{gman}). Performance visualization is handled by a customizable Logger object, which, apart from console logging, currently supports the Tensorboard and Vizdom backends.

\begin{figure}[!htb]
    \centering
    \includegraphics[width=\textwidth]{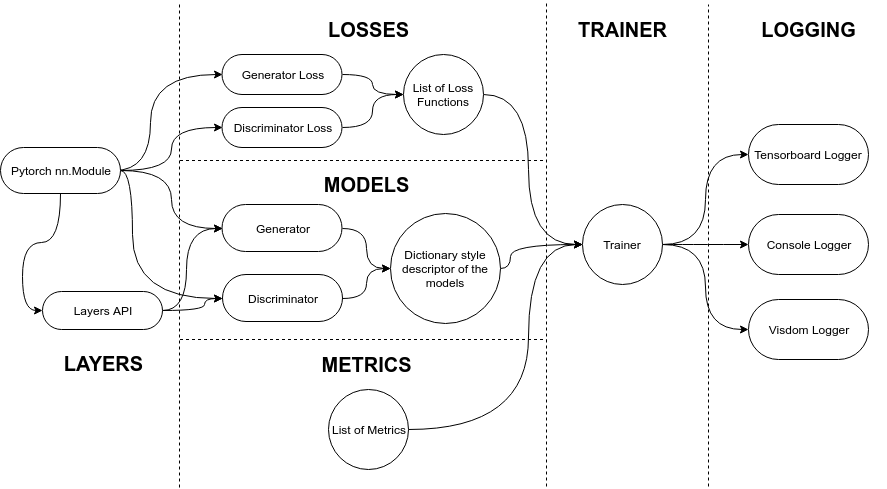}
    \caption{Overview of TorchGAN Design}
    \label{fig:tgan_design}
\end{figure}

\section{Comparison with Existing Frameworks}
\label{sec:frameworks}

TorchGAN provides high-quality implementations of various GAN models, metrics for evaluating GANs, and various approaches for improving the stability of GAN training. We provide an overview of the features that are provided off the shelf by TorchGAN and compare them with the ones provided by other frameworks. Note that the list is not exhaustive as the modular and extensible structure of TorchGAN allows one to extend or modify these features, or use them as building blocks for more sophisticated models.

\begin{center}
    \begin{tabular}{ c c c c c } 
        \toprule
        & TorchGAN & TFGAN & IBM GAN-Toolkit & HyperGAN\\
        \midrule
        Vanila GAN \cite{gan2014} & \checkmark & \checkmark & \checkmark & \checkmark\\
        DCGAN \cite{dcgan2015} & \checkmark & \checkmark & \checkmark & \checkmark\\
        Wasserstein GAN \cite{arjovsky2017wasserstein} & \checkmark & \checkmark & \checkmark & \checkmark\\
        Wasserstein GAN-GP \cite{gulrajani2017improved} & \checkmark & \checkmark & \checkmark & \checkmark\\
        Inception Score \cite{salimans2016improved} & \checkmark & \checkmark & \checkmark & \\
        InfoGAN \cite{chen2016infogan} & \checkmark & \checkmark & & \checkmark\\
        Cycle GAN \cite{cyclegan} & \checkmark & \checkmark & & \checkmark\\
        Least Squares GAN \cite{lsgan2017} & \checkmark & \checkmark & & \checkmark\\
        Auxillary Classifier GAN \cite{odena2017ac} & \checkmark & \checkmark & & \\
        Spectral Normalization GAN \cite{spectral2018} & \checkmark & \checkmark & & \\
        Self Attention GAN \cite{selfattn2018} & \checkmark & \checkmark & & \\
        Conditional GAN \cite{mirza2014conditional} & \checkmark & & \checkmark & \\
        Energy Based GAN \cite{zhao2016energybased} & \checkmark & & & \checkmark\\
        Boundary Equilibrium GAN \cite{berthelot2017began} & \checkmark & & & \\
        DRAGAN-GP \cite{kodali2017convergence} & \checkmark & & & \\
        Binary GAN \cite{dong2018training}& \checkmark & & & \\
        \makecell{Generative Multi Adversarial \\Networks (GMAN)} \cite{gman} & \checkmark & & & \\
        Adversarial Autoencoders \cite{zhang2017adversarial} & \checkmark & & & \\
        Historical Averaging \cite{salimans2016improved} & \checkmark & & & \\
        Feature Matching \cite{salimans2016improved} & \checkmark & & & \\
        Minibatch Discrimination \cite{salimans2016improved} & \checkmark & & & \\
        Frechet Inception Distance \cite{brock2018large} & $\star$ & \checkmark & \checkmark & \\
        Progressive GAN \cite{karras2017progressive} & $\star$ & \checkmark & & \\
        Adversarially Learned Inference  \cite{dumoulin2016adversarially} & $\star$ & & & \checkmark \\
        Star GAN \cite{Choi_2018} & & \checkmark & & \\
        \bottomrule
    \end{tabular}
    \captionof{table}{Supported features of different frameworks. Features officially supported are marked "\checkmark", under active development are marked "$\star$" and those currently unsupported are left blank.}
    \label{tab:frameworks}
\end{center}

Table~\ref{tab:frameworks} summarizes the features supported by a variety of open-source GAN frameworks. It suggests that TorchGAN supports the widest variety of features among the frameworks being considered. For comparison, we only consider the models present in the official repository of a given framework or an associated officially maintained model-zoo/examples repository. We avoid comparisons with projects like Pytorch-GAN\footnote{https://github.com/eriklindernoren/PyTorch-GAN}, Keras-GAN\footnote{https://github.com/eriklindernoren/Keras-GAN}, etc., as these are not frameworks and hence cannot be extended to newer models.

\section{Performance Benchmarks}

In order to demonstrate that TorchGAN incurs zero training overhead despite the high level of abstraction it provides, we compare the training time of TorchGAN with vanilla PyTorch implementations of DCGAN (\cite{dcgan2015}), CGAN (\cite{mirza2014conditional}), BEGAN (\cite{berthelot2017began}) and WGAN-GP (\cite{gulrajani2017improved}). Table~\ref{tab:benchmarks} reports the training time for TorchGAN and Pytorch for 1 epoch, averaged over 8 runs.

\begin{center}
    \begin{tabular}{ c c c c c }
        \toprule
        & \textbf{DCGAN} & \textbf{CGAN} & \textbf{WGAN-GP} & \textbf{BEGAN} \\
        \midrule
        & \scriptsize{\textit{CIFAR-10}} & \scriptsize{\textit{MNIST}} & \scriptsize{\textit{MNIST}} & \scriptsize{\textit{MNIST}} \\
        \midrule
        \textbf{TorchGAN} & \textbf{15.9s $\pm$ 0.64s} & \textbf{21.8s $\pm$ 0.43s} & \textbf{30.6s $\pm$ 1.35s} & \textbf{86.0s $\pm$ 0.62s} \\
        \textbf{Pytorch}  & 16.7s $\pm$ 0.24s & 22.4s $\pm$ 0.52s & 31.1s $\pm$ 0.97s & 87.0s $\pm$ 0.27s \\
        \bottomrule
    \end{tabular}
    \captionof{table}{Average Training Time : TorchGAN vs Pytorch Baselines}
    \label{tab:benchmarks}
\end{center}

For a fair comparison, we disable any form of logging and compute the training time using the $\%timeit$ magic function. We train the models on the CIFAR10 (\cite{Krizhevsky2009LearningML}) and MNIST\footnote{http://yann.lecun.com/exdb/mnist/} datasets, with a batch size of 128, on an Nvidia GTX Titan X GPU. 

\section{Development}

The source code is maintained at \href{https://github.com/torchgan/torchgan}{https://github.com/torchgan/torchgan} and is available under the MIT License. The package is lightweight and easy to install, with dependencies only on numpy, pytorch, and torchvision. An extensive set of examples are present in the main repository
(\href{https://github.com/torchgan/torchgan/tree/master/examples}{https://github.com/torchgan/torchgan}), documentation (\href{https://torchgan.readthedocs.io}{https://torchgan.readthedocs.io}) and model zoo (\href{https://github.com/torchgan/model-zoo}{https://github.com/torchgan/model-zoo}). To uphold quality and maintainability, we follow a strict code review process with a comprehensive test suite and continuous integration services.

\section{Conclusion and Future Work}

We present the features of the TorchGAN framework and demonstrate its extensibility, ease of use and efficiency.  Future work and extensions under active development include, integration of GAN models for video generation, generalization of the training loop to support Inference GAN models, such that they can be conveniently modified and extended, addition of features such as Adaptive Instance Normalization layers, and expanding the model zoo and documentation to cover more sophisticated examples such as Multi Agent-GAN training. We also envision the extension of the framework to domains beyond Computer Vision by adding support for NLP and Time Series GAN models.

\bibliography{ref}
\bibliographystyle{unsrt}

\end{document}